\documentclass{article}

\usepackage{microtype}
\usepackage{graphicx}
\usepackage{subfigure}
\usepackage{booktabs} 
\usepackage{header}

\usepackage{hyperref}

\usepackage[accepted]{icml2019}

\icmltitlerunning{Counterfactual Off-Policy Evaluation with Gumbel-Max SCMs}

\begin{document}
\twocolumn[
\icmltitle{Counterfactual Off-Policy Evaluation with \\
Gumbel-Max Structural Causal Models}


\icmlsetsymbol{equal}{*}

\begin{icmlauthorlist}
\icmlauthor{Michael Oberst}{mit}
\icmlauthor{David Sontag}{mit}
\end{icmlauthorlist}

\icmlaffiliation{mit}{CSAIL, Massachusetts Institute of Technology, Cambridge, MA, USA}

\icmlcorrespondingauthor{Michael Oberst}{moberst@mit.edu}

\icmlkeywords{Machine Learning, ICML, Causal Inference, Structural Causal Models, Reinforcement Learning, Off-Policy Evaluation}

\vskip 0.3in
]%



\printAffiliationsAndNotice{}  

\begin{abstract}
We introduce an off-policy evaluation procedure for highlighting episodes where applying a reinforcement learned (RL) policy is likely to have produced a substantially different outcome than the observed policy.  In particular, we introduce a class of structural causal models (SCMs) for generating counterfactual trajectories in finite partially observable Markov Decision Processes (POMDPs).  We see this as a useful procedure for off-policy ``debugging'' in high-risk settings (e.g., healthcare); by decomposing the expected difference in reward between the RL and observed policy into specific episodes, we can identify episodes where the counterfactual difference in reward is most dramatic.  This in turn can be used to facilitate review of specific episodes by domain experts. We demonstrate the utility of this procedure with a synthetic environment of sepsis management.
\end{abstract}

\section{Introduction}%
\label{sec:introduction}

When a patient dies in the hospital, we might ask: Could it have been avoided, if the clinicians acted differently?  This is impossible to know, because we cannot go back in time.  Nonetheless, there is precedent for trying to answer these counterfactual questions: In medical malpractice, for instance, establishing fault requires showing that the injury would not have occurred ``but for'' the breach in the standard of care~\citep{Bal2009, VariousAuthors2008}.

When considering the deployment of new technologies in high-risk environments, such as RL policies in healthcare, we might worry about a similar counterfactual:  If we had deployed an RL policy, would it have caused the death of patients who lived under a physician policy?  Or more optimistically, are there lives which could have been saved?  This question becomes increasingly relevant, with growing interest in applying RL to everything from HIV therapy \citep{Parbhoo2017} to epilepsy \citep{Guez2008} and sepsis \citep{Komorowski2018}.

However, evaluation of RL policies with observational data remains a challenge, albeit a necessary one due to the safety risk of training an RL agent on real patients.  Off-policy evaluation in this context is prone to issues such as confounding, small effective sample sizes, and lack of introspection as discussed in \citet{Gottesman2019}.  Lacking a surefire solution, a physician might review patient trajectories alongside the suggested actions from an RL policy, and ask: Does this seem reasonable?  What do I think would have happened if this policy had been followed?  Yet, manual inspection of trajectories is not only inefficient, but difficult to interpret without more information: If we are to discover new insights about treatment, shouldn't there be \textit{some} disagreement?

This motivates our main \textit{conceptual} contribution: By flipping the counterfactual question around, and ``asking the algorithm'' what would have happened, we aim to highlight the most interesting cases of disagreement, while also demonstrating implicitly why the learned policy is preferred, based on the projected differences in trajectories.  

More specifically, in the context of model-based RL with discrete states, we give a post-hoc method to elicit the answers to these questions based on a learned dynamics model.  Our goal is to pair counterfactual trajectories with observed trajectories, so that domain experts can ``sanity-check'' a proposed policy, potentially with additional side-information (e.g., chart review in the case of a patient).  In Section~\ref{sec:experiments} we use a synthetic environment of sepsis management to demonstrate how this type of introspection can highlight dangerous implicit assumptions of the learned policy, even when off-policy evaluation is overly optimistic.

Towards generating these counterfactual trajectories, we have to deal with a fundamental issue of non-identifiability in the absence of deterministic dynamics.  As we show in this paper, even with a fully-specified finite Markov Decision Process (MDP), there are multiple parameterizations of a structural causal model (SCM) which are equally compatible with the transition and reward distributions, but which suggest different counterfactual outcomes.  For binary treatment and outcomes, the \textit{monotonicity} condition is sufficient to identify the counterfactual distribution \citep{Pearl2000} and has been previously used in epidemiology applications \citep{Cuellar2018}. However, to our knowledge, there is no analogous condition for the categorical case.

This motivates our main \textit{theoretical} contribution, which is two-fold.  First, we introduce a general condition of \textit{counterfactual stability} for SCMs with categorical variables (e.g., state transitions) and prove that this condition implies the monotonicity condition in the case of binary categories.  Second, we introduce the \textit{Gumbel-Max SCM}, based on the Gumbel-Max trick for sampling from discrete distributions, and (a) demonstrate that it satisfies the counterfactual stability condition, and (b) give a Monte Carlo procedure for drawing counterfactual trajectories under this model.  We note that any discrete probability distribution can be sampled using a Gumbel-Max SCM\@; As a result, drawing counterfactual trajectories can be done in a post-hoc fashion, given any probabilistic model of dynamics with discrete states.

The paper proceeds as follows: In Section~\ref{sec:preliminaries}, we review the formulation of structural causal models and counterfactual estimation, as given by \citet{Pearl2009} and \citet{Peters2017}.  In Section~\ref{sec:theoretical_contributions}, we introduce the assumption of counterfactual stability, propose a SCM which satisfies it, and show how it can be used to estimate counterfactual trajectories in a Partially Observed Markov Decision Process (POMDP).  In Section~\ref{sec:related_work}, we discuss related work, and in Section~\ref{sec:experiments} we give an illustrative application using a synthetic example.

\section{Preliminaries}%
\label{sec:preliminaries}

\subsection{Structural Causal Models}%
\label{sub:structural_causal_models}

First, we briefly review the concept of \textit{structural causal models}, and encourage the reader to refer to \citet{Pearl2009} (Section 7.1) and \citet{Peters2017} for more details.  

\textbf{Notation:} As a general rule throughout, we refer to a random variable with a capital letter (e.g., $X$), the value it obtains as a lowercase letter (e.g., $X = x$), and a set of random variables with boldface font (e.g., $\mathbf{X} = \{X_1, \ldots, X_n\}$).  Consistent with \citet{Peters2017} and \citet{Buesing2018}, we write $P_X$ for the distribution of a variable $X$, and $p_x$ for the density function.

\begin{thmdef}[Structural Causal Model (SCM)]
  A structural causal model $\cM$ consists of a set of independent random variables $\bU = \{U_1, \ldots, U_n\}$ with distribution $P(\bU)$, a set of functions $\mathbf{F} = \{f_1, \ldots, f_n\}$, and a set of variables $\bX = \{X_1, \ldots, X_n\}$ such that $X_i = f_i(\mathbf{PA}_i, U_i), \forall i$, where $\mathbf{PA}_i \subseteq \bX \setminus X_i$ is the subset of $\bX$ which are parents of $X_i$ in the causal DAG $\cG$.  As a result, the prior distribution $P(\bU)$ and functions $\mathbf{F}$ determine the distribution $P^\cM$.
\end{thmdef}

\begin{figure}
\centering
  \begin{subfigure}
    \centering
    \begin{tikzpicture}[
      obs/.style={circle, draw=gray!50, fill=gray!30, very thick, minimum size=8mm},
      uobs/.style={circle, draw=gray!50, fill=white, very thick, minimum size=8mm},
      calc/.style={draw=white, text=white, fill=black, very thick, minimum size=8mm},
      ]
      \node[obs] (Y) {$Y$};
      \node[obs] (T) [left=0.5cm of Y] {$T$};
      \node[obs] (X) [above=0.5cm of T] {$\bX$};
      \draw[-latex, thick] (X) -- (Y);
      \draw[-latex, thick] (X) -- (T);
      \draw[-latex, thick] (T) -- (Y);
    \end{tikzpicture}%
  \end{subfigure}
  \begin{subfigure}
    \centering
    \begin{tikzpicture}[
      obs/.style={circle, draw=gray!50, fill=gray!30, very thick, minimum size=8mm},
      uobs/.style={circle, draw=gray!50, fill=white, very thick, minimum size=8mm},
      calc/.style={draw=white, text=white, fill=black, very thick, minimum size=8mm},
      ]
      \node[calc] (Y) {$Y$};
      \node[calc] (T) [left=0.5cm of Y] {$T$};
      \node[calc] (X) [above=0.5cm of T] {$\bX$};
      \node[uobs] (Ut) [left=0.5cm of T] {$U_t$} ;
      \node[uobs] (Ux) [left=0.5cm of X] {$U_x$} ;
      \node[uobs] (Uy) [above=0.5cm of Y] {$U_y$} ;
      \draw[-latex, thick] (X) -- (Y);
      \draw[-latex, thick] (X) -- (T);
      \draw[-latex, thick] (T) -- (Y);
      \draw[-latex, thick] (Uy) -- (Y);
      \draw[-latex, thick] (Ux) -- (X);
      \draw[-latex, thick] (Ut) -- (T);
    \end{tikzpicture}%
  \end{subfigure}
  \caption{Example translation of a Bayesian Network into the corresponding Structural Causal Model.  \textbf{Left:} Causal DAG on an outcome $Y$, covariates $X$, and treatment $T$.  Given this graph, we can perform do-calculus \citep{Pearl2009} to estimate the impact of interventions such as $\E[Y | X, do(T = 1)] - \E[Y | X, do(T = 0)]$, known as the Conditional Average Treatment Effect (CATE).  \textbf{Right:} All observed random variable are assumed to be generated via structural mechanisms $f_x, f_t, f_y$ via independent latent factors $U$ which cannot be impacted via interventions.  Following convention of \citet{Buesing2018}, calculated values are given by black boxes (and in this case, are observed), observed latent variables are given in grey, and unobserved variables in white.}%
  \label{fig:cate_dag}
\end{figure}
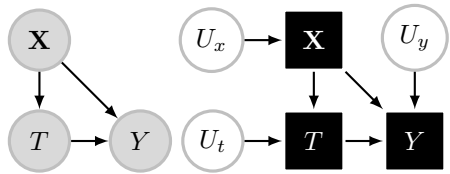

As a motivating example to simplify exposition, we will assume the pair of causal graphs given in Figure~\ref{fig:cate_dag}.  This causal graph corresponds to a single-action setting which is often assumed in the literature of Conditional Average Treatment Effect (CATE) estimation \citep[see, e.g.,][]{Johansson2016LearningInference}.  It can be used to represent, for instance, the effect of a medical treatment $T$ on an outcome $Y$ in the presence of confounding variables $\bX$.

The SCM $\cM$ defines a complete data-generating processes, which entails the \textit{observational distribution} $P(\bX, Y, T)$.  It also defines \textit{interventional distributions}.  For example, the estimate of CATE is given by $\tau_x = \E[Y | X, do(T = 1)] - \E[Y | X, do(T = 0)]$, where the \textit{do-operator} \citep{Pearl2009} is used to signify an intervention. 

\begin{thmdef}[Interventional Distribution]
  Given an SCM $\cM$, an intervention $I = do\left(X_i \coloneqq \tilde{f}(\tilde{\mathbf{PA}}_i, \tilde{U}_i)\right)$ corresponds to replacing the structural mechanism $f_i(\mathbf{PA}_i, U_i)$ with $\tilde{f}_i(\tilde{\mathbf{PA}}_i, U_i)$.  This includes the concept of atomic interventions, where we may write more simply $do(X_i = x)$.  The resulting SCM is denoted $\cM^I$, and the resulting interventional distribution is denoted $P^{\cM; I}$.
\end{thmdef}

For instance, suppose that $Y$ corresponds to a favorable binary outcome, such as 5-year survival, and $T$ corresponds to a treatment. Then several quantities of interest in causal effect estimation, such as $p^{T = 1} \coloneqq \E[Y | X, do(T = 1)]$, $p^{T = 0} \coloneqq \E[Y | X, do(T = 0)]$, and $\tau_x = p^{T = 1} - p^{T = 0}$, are all defined by the interventional distribution, which is \textit{forward-looking}, telling us what might be expected to occur if we applied an intervention.  

However, we can also define the \textit{counterfactual distribution} which is \textit{retrospective}, telling us what might have happened had we acted differently.  For instance, we might ask:  Having given the drug and observed that $Y = 1$ (survival), what \textit{would have happened} if we had instead withheld the drug?

\begin{thmdef}[Counterfactual Distribution]
  Given an SCM $\cM$ and an observed assignment $\bX = \bx$ over any set of observed variables, the counterfactual distribution $P^{\cM|\bX = \bx; I}$ corresponds to the distribution entailed by the SCM $\cM^I$ using the posterior distribution $P(\bU | \bX = \bx)$.
\end{thmdef}

Explicitly, given an SCM $\cM$, the counterfactual distribution can be estimated by first inferring the posterior over latent variables, e.g., $P(\bU|\bX = \bx, T = 1, Y = 1)$ in our running example, and then passing that distribution through the structural mechanisms in a modified $\cM^I$ (e.g., $I = do(T = 0)$) to obtain a counterfactual distribution over any variable\footnote{Called abduction, action, and prediction in \citet{Pearl2009}}.

\subsection{(Non)-Identifiability of Counterfactuals}%
\label{sub:identifiability_of_counterfactuals}

Given an SCM $\cM$, we can compute an answer to our counterfactual question:  Having given the drug and observed that $Y = 1$ (survival), what \textit{would have happened} if we had instead withheld the drug?  In the binary case, this corresponds to the \textit{Probability of Necessity} (PN) \citep{Pearl2009, Dawid2015}, because it represents the probability that the exposure $T = 1$ was necessary for the outcome.

However, the answer to this question is not identifiable without further assumptions:  That is, there are multiple SCMs which are all consistent with the interventional distribution, but which produce different counterfactual estimates of quantities like the Probability of Necessity \citep{Pearl2009}.  

\subsection{Monotonicity Assumption to Identify Binary Counterfactuals}%
\label{sub:assumptions_to_identify_counterfactuals}

Nonetheless, there are plausible (though untestable) assumptions we can make that identify counterfactual distributions.  In particular, the \textit{monotonicity assumption} \citep{Pearl2000, Tian2000} is sufficient to identify the Probability of Necessity and related quantities used in epidemiology \citep{Cuellar2018, Yamada2017} to answer counterfactual queries with respect to a binary treatment and outcome. 
\begin{thmdef}[Monotonicity]
  An SCM of a binary variable $Y$ is monotonic relative to a binary variable $T$ if and only if it has the following property\footnote{We could also write this property as conditional on $X$}\textsuperscript{,}\footnote{This definition differs slightly from the presentation of monotonicity in \citet{Pearl2009}, where $f_y(t, u)$ being monotonically increasing in $t$ is given as the property, with the testable implication that $\E[Y | do(T = t)] \geq \E[Y | do(T = t')]$ for $t \geq t'$. Because the direction of monotonicity is only compatible with the corresponding direction of the expected interventional outcomes, we fold this into the definition of monotonicity directly, to align with our later definition of counterfactual stability. Also note that we use the notation $Y_{do(T = t)} \coloneqq f_y(t, u)$ here}: $\E[Y | do(T = t)] \geq \E[Y | do(T = t')] \implies f_y(t, u) \geq f_y(t', u), \ \forall u$.  We can write equivalently that the following event never occurs, in the case where $\E[Y | do(T = 1)] \geq \E[Y | do(T = 0)]$: $Y_{do(T = 1)} = 0 \land Y_{do(T = 0)} = 1$.  Conversely for $\E[Y | do(T = 1)] \leq \E[Y | do(T = 0)]$, the following event never occurs: $Y_{do(T = 1)} = 1 \land Y_{do(T = 0)} = 0$.
  \label{def:monotonicity}
\end{thmdef}
  
 This assumption restricts the class of possible SCMs to those which all yield equivalent counterfactual distributions over $Y$.  For instance, the following SCM exhibits the monotonicity property, and replicates any interventional distribution where $g(x, t) = \E[Y | X = x, do(T = t)]$:
\[
  Y = \1{U_y \leq g(x, t)}, \quad U \sim \textrm{Unif}(0, 1)
\]
More importantly, the monotonicity assumption identifies several counterfactual quantities, such as the Probability of Necessity mentioned earlier.

\subsection{Relationship to POMDPs}%
\label{sub:relationship_to_pomdps}

\begin{figure}
  \centering
  \includegraphics[width=0.95\linewidth]{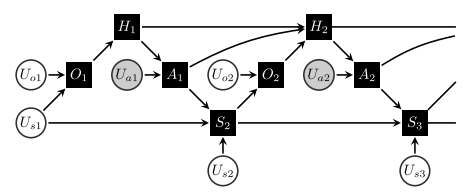}
  \caption{SCM for a POMDP as taken from \citet{Buesing2018}, with initial state $U_{s1} = S_1$, states $S_t$, and histories $H_t$, where the action is generated via the mechanism $\pi(U_a, H_t)$.}%
  \label{fig:pomdp_graph}
\end{figure}

As noted in \citet{Buesing2018}, we can view an episodic Partially Observable Markov Decision Process (POMDP) as an SCM, as shown in Figure~\ref{fig:pomdp_graph}, where $S_t$ corresponds to states, $A_t$ corresponds to actions, $O_t$ corresponds to observable quantities (including reward $R_t$), $H_t$ contains history up to time $t$, i.e., $H_t = \{O_1, A_1, \ldots A_{t-1}, O_t\}$, and stochastic policies are given by $\pi(a_t | h_t)$.  

\citet{Buesing2018} investigate a grid-world setting where the transition distribution $P(S_{t+1} | S_{t}, A_t)$ is deterministic.  To extend this counterfactual reasoning to stochastic transitions, we need to make further assumptions on the functional relationships themselves.  However, to the best of our knowledge, there is no investigation of plausible assumptions in the literature to identify the answers to counterfactual queries in the case of categorical distributions.

\section{Gumbel-Max SCMs}%
\label{sec:theoretical_contributions}

In the context of reinforcement learning with POMDPs, we are typically concerned with estimating the expected reward of a proposed policy $\hat{\pi}$.  To formalize notation, a given policy $\pi$ implies a density over trajectories $\tau \in \cT = (S_1, O_1, A_1, \ldots, A_{T-1}, S_T, O_T)$, which we denote as $p^{\pi}(\tau)$, and we let $R(\tau)$ be the total reward of a trajectory $\tau$. For ease of notation, we sometimes write $\E_{\hat{\pi}}$ and $\E_{obs}$ to indicate an expectation taken with respect to $\tau \sim p^{\hat{\pi}}$ and $\tau \sim p^{\pi_{obs}}$ respectively, where $\hat{\pi}$ refers to the proposed (`target') policy, and $\pi_{obs}$ to the observed (`behavior') policy.

If we wish to compare the performance of a proposed policy $\hat{\pi}$ and the observed policy $\pi_{obs}$, we might compare the difference in expected reward.  The expected reward under $\pi_{obs}$ can be estimated in this case using observed trajectories, without a model of the environment.  The difference in expected reward is conceptually similar to the average treatment effect (ATE) of applying the proposed vs.\ observed policy, and we denote it as $\delta$:
\begin{equation}
  \delta \coloneqq \E_{\hat{\pi}}[ R(\tau) ] - \E_{obs} [ R(\tau) ]
  \label{eq:expected_reward}
\end{equation}
However, it may be useful to drill down into specific cases: Perhaps there are certain environments, for instance, in which the proposed policy would perform better or worse than the observed policy.  One natural decomposition is to condition on the first observed state to estimate a conditional expected reward, e.g.,
\begin{equation}
  \delta_{o} \coloneqq \E_{\hat{\pi}}[ R(\tau) | O_1 = o ] - \E_{obs} [ R(\tau) | O_1 = o ]
  \label{eq:expected_conditional_reward}
\end{equation}
Equation~\ref{eq:expected_conditional_reward} corresponds conceptually to CATE estimation, where we condition only on pre-treatment information (in this case, $O_1$, which occurs before the first action).  We can use post-treatment information to decompose this further, over actual trajectories that we have observed, to highlight differences between the observed and proposed policy.  Given a causal model of the environment (in the form of an SCM as described in Section~\ref{sub:relationship_to_pomdps}), we can decompose Equation~\ref{eq:expected_conditional_reward} further as follows:
\begin{thmlem}[Counterfactual Decomposition of Expected Reward]
  Let trajectories $\tau$ be drawn from $p^{\pi_{obs}}$.  Let $\tau_{\hat{\pi}}$ be a counterfactual trajectory, drawn from our posterior distribution over the exogenous $U$ variables under the new policy $\hat{\pi}$.  Note that under the SCM, $\tau_{\hat{\pi}}$ is a deterministic function of the exogenous $U$ variables, so we can write $\tau_{\hat{\pi}}(u)$ to be explicit:
\begin{align*}
  &\E_{\hat{\pi}}[ R(\tau) | O_1 = o ] \\
  &= \int_{\tau} p^{\pi_{obs}}(\tau | O_1 = o_1) \E_{u \sim p^{\pi_{obs}}(u | \tau)}[R(\tau_{\hat{\pi}}(u))] d\tau
\end{align*}
  \label{thm:decomposition}
\end{thmlem}
\begin{thmcol}[Counterfactual Decomposition of $\delta_o$]
\begin{align*}
  \delta_{o} &\coloneqq \E_{\hat{\pi}}[ R(\tau) | O_1 = o ] - \E_{obs} [ R(\tau) | O_1 = o ] \\
             &= \int_{\tau} p^{\pi_{obs}}(\tau | O_1 = o_1) \E_{u \sim p^{\pi_{obs}}(u | \tau)}[R(\tau_{\hat{\pi}}(u)) - R(\tau)]d\tau
\end{align*}
\label{thm:decomposition_col}
\end{thmcol}
\vspace{-4mm}
The proofs of Lemma~\ref{thm:decomposition} and Corollary~\ref{thm:decomposition_col} are very similar to (and are essentially implied by) Lemma 1 from \citet{Buesing2018}, but are given in self-contained form in the supplement.  Corollary~\ref{thm:decomposition_col} implies that we can decompose the expected difference in reward between the policies into differences on \textit{observed episodes} over counterfactual trajectories.  However, we face a non-identifiability issue when transitions are not deterministic:  Multiple SCMs can all entail the same interventional distribution, but a different set of counterfactual trajectories, and therefore a different decomposition under Lemma~\ref{thm:decomposition}.

This motivates the theoretical work of this section: We must make our assumptions carefully, as they cannot be tested by data, so it is worth investigating which assumptions are consistent with our causal intuition.  We illustrate this non-identifiability (with respect to categorical distributions) in Section~\ref{sec:nonid_categorical}.  Then we introduce the condition of \textit{counterfactual stability} (in Section~\ref{sec:counterfactual_stability}) for a discrete distribution on $k$ categories, and show that it is compatible with the monotonicity condition of \citet{Pearl2000} in that it implies the monotonicity assumption when $k=2$.  Then we introduce the Gumbel-Max SCM for discrete variables in Section~\ref{sec:gumbel_scm}, and prove that it satisfies the counterfactual stability condition.  Finally, in Section~\ref{sec:inference_with_gumbel} we describe how the Gumbel-Max SCM assumption can be used to generate a posterior over the counterfactual distribution, under the condition that the data comes from this SCM\@.

\subsection{Non-Identifiability of Categorical Counterfactuals}%
\label{sec:nonid_categorical}

We will first illustrate that the non-identifiability of counterfactual distributions applies to categorical distributions as well.  Consider the categorical distribution over $k$ categories, e.g., the transition kernel $P(S' | S = s, A = a)$ over discrete states.  Let $p_i \coloneqq P(S' = i | S = s, A = a)$.  There are multiple ways that we could sample from this distribution with a structural mechanism $f$ and latent variables $U$.  For instance, we could define an ordering $\textbf{ord}$ on the categories, and define $k$ intervals of $[0, 1]$ as $[0, p_{\textbf{ord}(1)}), [p_{\textbf{ord}(1)}, \sum_{i = 1}^{2} p_{\textbf{ord}(i)}), \ldots, [\sum_{i = 1}^{k-1} p_{\textbf{ord}(i)}, 1]$.  Then we could draw $U \sim Unif(0, 1)$, and return the interval that $u$ falls into.  

However, different permutations $\textbf{ord}$ will yield equivalent interventional distributions but can imply different counterfactual distributions.  For instance, let $k=4$ and $p_1 = p_2 = 0.25, p_3 = 0.3, p_4 = 0.2$ and consider an intervention $A = a'$ which defines a different distribution $p'_1 = 0, p'_2 = 0.25, p'_3 = 0.25, p'_4 = 0.5$.  Now consider two permutations, $\textbf{ord} = [1,2,3,4]$ and $\textbf{ord}' = [1,2,4,3]$, and the counterfactual distribution under $a'$ given that $S' = 2, A = a$.  In each case, posterior inference over $U$ implies that $P(U | S' = 2, S = s, A = a) \sim Unif[0.25, 0.5)$. However, under $\textbf{ord}$ this implies the counterfactual $S' = 3$, while under $\textbf{ord}'$ it implies $S' = 4$. A visual depiction of this can be found in the supplement.

Note that in this example, the mechanism $f_\mathbf{ord}$ implied a non-intuitive counterfactual outcome:  Even though the intervention $A = a'$ \textit{lowered} the probability of $S' = 3$ (relative to the probability under $A = a$) without modifying the probability of $S' = 2$, it led to a counterfactual outcome of $S' = 3$.  Since all choices for $\mathbf{ord}$ imply the same interventional distribution, there is no way to distinguish between these mechanisms with data.

This motivates the following sections, where we posit a desirable property for categorical SCMs which rules out this result (among others) and is compatible with the notion of monotonicity introduced by \citet{Pearl2000}.  We then demonstrate that a mechanism based on sampling independent Gumbel variables satisfies this property, which motivates the use of the Gumbel-Max SCM for performing counterfactual inference in this setting. 

\subsection{Counterfactual Stability}%
\label{sec:counterfactual_stability}

We now introduce our first contribution, the desired property of \textit{counterfactual stability} for categorical SCMs with $k$ categories, laid out in in Definition~\ref{def:counterfactual_stability}.  This property would rule out the non-intuitive counterfactual implications of $f_\mathbf{ord}$ in Section~\ref{sec:nonid_categorical}.  We then demonstrate that this condition implies the monotonicity condition when $k=2$.  

\textbf{Notation:} Denote the interventional probability distribution of a categorical variable $Y$ with $k$ categories as $P^{\cM; I}(Y) = \mathbf{p}$ under intervention $I$, and $\mathbf{p}'$ under intervention $I'$, where $\mathbf{p}, \mathbf{p}'$ are vectors in the probability simplex over $k$ categories.  To simplify notation for interventional outcomes, we will sometimes denote by $Y_{I}$ the observed outcome $Y$ under intervention $I$, and denote by $Y_{I'}$ the counterfactual outcome under intervention $I'$, such that $p_i$ and $P(Y_{I} = i)$ are both equivalent to $P^{\cM; I}(Y = i)$, and similarly for $I'$.  For counterfactual outcomes, we will write $P^{\cM| Y_I = i; I'}(Y)$ for the counterfactual distribution of $Y$ under intervention $I'$ given that we observed $Y = i$ under the intervention $I$.

\begin{thmdef}[Counterfactual Stability]
  An SCM of a categorical variable $Y$ satisfies \textit{counterfactual stability} if it has the following property:  If we observe $Y_{I} = i$, then for all $j \neq i$, the condition $\frac{p'_i}{p_i} \geq \frac{p'_j}{p_j}$ implies that $P^{\cM|Y_I=i; I'}(Y = j) = 0$.  That is, if we observed $Y = i$ under intervention $I$, then the counterfactual outcome under $I'$ cannot be equal to $Y = j$ unless the multiplicative change in $p_i$ is less than the multiplicative change in $p_j$.
  \label{def:counterfactual_stability}
\end{thmdef}
\begin{thmcol}
  If $\cM$ is a SCM which satisfies counterfactual stability, then if we observe $Y_{I} = i$, and $\frac{p'_i}{p_i} \geq \frac{p'_j}{p_j}$ holds for all $j \neq i$, then $P^{\cM| Y_I=i; I'}(Y = i) = 1$.%
  \label{thm:same_outcome}
\end{thmcol}

This definition and corollary encode the following intuition about counterfactuals:  If we had taken an alternative action that would have \textit{only increased} the probability of $Y = i$, without increasing the likelihood of other outcomes, then the same outcome would have occurred in the counterfactual case.  Moreover, in order for the outcome to be different under the counterfactual distribution, the relative likelihood of an alternative outcome must have increased relative to that of the observed outcome.  The connection to monotonicity is given in Theorem~\ref{thm:cf_stable_is_monotonic}.  The proof is straightforward, and found in the supplement.

\begin{thmthm}
  Let $Y = f_y(t, u)$ be the SCM for a binary variable $Y$, where $T$ is also a binary variable.  If this SCM satisfies the counterfactual stability property, then it also satisfies the monotonicity property with respect to $T$.
  \label{thm:cf_stable_is_monotonic} 
\end{thmthm}

\subsection{Gumbel-Max SCM}%
\label{sec:gumbel_scm}

Unlike monotonicity with binary outcomes and treatments, the condition of counterfactual stability does not obviously imply any closed-form solution for the counterfactual posterior.  Thus, we introduce a specific SCM which satisfies this property, and discuss in Section~\ref{sec:inference_with_gumbel} how to sample from the posterior distribution in a straightforward fashion.

We recall the following fact, known as the Gumbel-Max trick \citep{Luce1959, Yellott1977, Yuille.2011, Hazan2012, Maddison2014, Hazan2016, Maddison2017}: 
\begin{thmdef}[Gumbel-Max Trick]
  We can sample from a categorical distribution with $k$ categories as follows, where $\tilde{p}_i$ is the unnormalized probability $P(Y = i)$:  First, draw $g_1, \ldots, g_k$ from a standard Gumbel, which can be achieved by drawing $u_1, \ldots, u_k$ iid from a Unif$(0, 1)$, and assigning $g_i = - \log (-\log u_i)$. Then, set the outcome $j$ by taking $\argmax_j \log \tilde{p}_j + g_j$.%
\label{def:gumbel_max_trick}
\end{thmdef}
Clearly, we can perform this for any categorical distribution, e.g., the transition distribution $p_i = P(S' = i | S, A)$;  In particular, for any discrete variable $Y$ whose parents in a causal DAG are denoted $\bX$, a \textit{Gumbel-Max SCM} assumes the following causal mechanism, where $\mathbf{g} = (g_1, \ldots, g_k)$ are independent Gumbel variables:
\begin{equation*}
  Y = f_y(\bx, \mathbf{g}) \coloneqq \argmax_{j}\{ \log P(Y = j | \mathbf{X} = \mathbf{x}) + g_j \}
\end{equation*}
Like any mechanism which replicates the conditional distribution under intervention, this mechanism is indistinguishable from any other causal mechanism based on data alone.  That said, it does satisfy the property given in Definition~\ref{def:counterfactual_stability}.  \begin{thmthm}
  The Gumbel-Max SCM satisfies the counterfactual stability condition.%
  \label{thm:gumbel_counterfactual_stability}
\end{thmthm}
The proof is straightforward, and given in the supplement.  The intuition is that, when we consider the counterfactual distribution, the Gumbel variables are fixed.  Thus, in order for the argmax (our observed outcome) to change in the counterfactual, the log-likelihood of an alternative outcome must have increased relative to our observed outcome.

\subsection{Posterior Inference in the Gumbel-Max SCM}%
\label{sec:inference_with_gumbel}

Given a Gumbel-Max SCM as defined above, where $Y = \argmax_{j} \log p_j + g_j$ and $p_j \coloneqq P(Y_I = j)$, we can draw Monte Carlo samples from the posterior $P(\mathbf{g} | Y_I = i)$ using one of two approaches:  First, we can use rejection sampling, drawing samples from the prior $P(\mathbf{g})$ and rejecting those where $i \neq \argmax_j \log p_j + g_j$.  Alternatively, it is known \citep{Maddison2014, Maddison2017a} that in the posterior, the maximum value and the argmax of the shifted Gumbel variables $\log p_j + g_j$ are independent, and the maximum value is distributed as a standard Gumbel (in the case of normalized probabilities).  Thus, we can sample the maximum value first, and then sample the remaining values from shifted Gumbel distributions that are truncated at this maximum value.  Then, for each index $j$, subtracting off the location parameter $\log p_j$ will give us a sample of $g_j$. We can then add this sample $\mathbf{g}$ to the log-probabilities under $I'$ (i.e., $\log \mathbf{p}'$) and take the new argmax to get a sample of the counterfactual outcome $Y$ under intervention $I'$.

\section{Related Work}%
\label{sec:related_work}

In the machine learning community, there has been recent interest in applying insights from causal inference to augment the training of RL models, such as in bandit settings \citep{NIPS2018_7523} and in model-based RL \citep{Buesing2018}.  The most similar work to our own is \citet{Buesing2018}, but while they use counterfactuals to approximate draws from the interventional distribution, we treat the counterfactual distribution as the primary object of interest.

The term `counterfactuals' is often used more broadly in causal inference and machine learning, e.g., in the estimation of potential outcomes \citep{Imbens2015}, which are also counterfactual outcomes at an individual level.  However, this is often used \citep[e.g.,][]{Schulam2017, Johansson2016LearningInference} primarily towards predicting quantities related to the interventional distribution, such as CATE.

Assumptions on structural mechanisms are also used in structure learning to identify the direction of causation $X \rightarrow Y$ or $X \leftarrow Y$ from the observational distribution alone, with both continuous \citep{Peters2014, Mooij2016} and discrete \citep{Kocaoglu2017} variables.  However, these assumptions imply differences in the observational distribution (by design), whereas our assumptions distinguish between counterfactual distributions which are both observationally and interventionally equivalent.  

The assumption of monotonicity implicitly appears in early work in epidemiology on estimating quantities like the `relative risk ratio' \citep{Miettinen1974}, which are often imbued with causal interpretations \citep{Pearl2009, Yamada2017}. Formalizing the assumption of monotonicity, required to correctly impute causal meaning to these quantities, is covered in \citet{Balke1994, Pearl2000, Tian2000}.  More recent work in epidemiology uses the assumption of monotonicity explicitly, e.g., to estimate the counterfactual effect of water sanitation in Kenya \citep{Cuellar2018}, and there has been ample discussion and debate regarding how this reasoning could apply (in principle) to legal cases, such as litigation around the toxic effects of drugs \citep{PhilipDawid2016}.

The use of Gumbel variables to sample from a categorical distribution (the Gumbel-Max Trick) is well known (see Definition~\ref{def:gumbel_max_trick}) but in recent work in the machine learning community \citep{Maddison2017, Hazan2016} it is not imbued with the same causal / counterfactual interpretation that we propose here.  The Gumbel-Max mechanism was initially introduced in the discrete-choice literature \citep{Luce1959}, where it is used as a generative model for decision-making under utility maximization \citep{Train2002, Aguirregabiria2010}, where the log probabilities may be assumed to follow some functional form, such as being linear in features. This is motivated by understanding the impact of different characteristics on consumer choices, see \citep[Example 1]{Aguirregabiria2010}.  In contrast, we decouple this structural mechanism (for estimation of counterfactuals) from the statistical model used to estimate the conditional probability distributions under interventions.

\section{Experiments}%
\label{sec:experiments}

As discussed in Section~\ref{sec:introduction}, our hope is to provide a method for qualitative introspection and `debugging' of RL models, in settings where a domain expert could plausibly examine individual trajectories.  We give an illustrative example of this use case here, motivated by recent work examining the use of RL algorithms for treating sepsis among intensive-care unit (ICU) patients \citep{Komorowski2018}.  In particular, we use a simple simulator of sepsis and ``debug'' an RL policy that is learned on observed trajectories.

An analysis like this requires three ingredients: First, we are given \textit{observed trajectories}, but cannot directly interact with the environment\footnote{We do not assume access to a simulator;  In this example, it is used only for obtaining the initial observed trajectories.}.  Second, we have access to a \textit{structural causal model} of the environment.  In this case, that model is a finite MDP, learned based on observed samples, combined with the assumption of a Gumbel-Max SCM for transition distributions.  Finally, we need a \textit{target policy} to evaluate.  We refer to the policy which generated the data as the \textit{behavior} policy, to distinguish it from the target policy.

In Sections~\ref{sec:illustrative_analysis_sepsis}-\ref{sec:off_policy_evaluation_is_misleading} we describe our illustrative scenario, in which a target RL policy appears to perform well using off-policy evaluation methods such as weighted importance sampling, when it is actually much worse than the behavior policy. In Sections~\ref{sec:id_informative_trajectories}-\ref{sec:insights_from_examining_individual_trajectories} we then demonstrate how our method could be used to identify a promising subset of trajectories for further introspection, and uncover the flaws in the target policy using side information (e.g., chart review of individual patients).  All the code required to reproduce our experiments is available online at \url{https://www.github.com/clinicalml/gumbel-max-scm}

\subsection{Illustrative Analysis: Sepsis}%
\label{sec:illustrative_analysis_sepsis}

\textbf{Environment:} Our simulator includes four vital signs (heart rate, blood pressure, oxygen concentration, and glucose levels) with discrete states (e.g., low, normal, high), along with three treatment options (antibiotics, vasopressors, and mechanical ventilation), all of which can be applied at each time step.  Reward is +1 for discharge of a patient, and -1 for death.  Discharge occurs only when all patient vitals are within normal ranges, and all treatments have been stopped.  Death occurs if at least three of the vital signs are simultaneously out of the normal range.  In addition, a binary variable for diabetes is present with 20\% probability, which increases the likelihood of fluctuating glucose levels. 

\textbf{Observed Trajectories}: For the purposes of this illustration, the behaviour policy was constructed using Policy Iteration \citep{Sutton2017} with full access to the parameters of the underlying MDP (including diabetes state). This was done deliberately to set up a situation in which the observed policy performs well. To introduce variation, the policy takes a random alternative action w.p. $0.05$. Using this policy, we draw 1000 patient trajectories from the simulator, with a maximum of 20 time steps. If neither death nor discharge is observed, the observed reward is zero.

\textbf{Structural Causal Model:} For this illustration, we `hide' glucose and diabetes state in the observed trajectories; Given this reduced state-space, we learn the parameters of the finite MDP by using empirical counts of transitions and rewards from the 1000 observed trajectories, with death and discharge treated as absorbing states.  For state / action pairs that are not observed, we assume that any action leads to death, and confirm that this results in a target policy which never takes an action that has never been observed.  For counterfactual evaluation, we make the assumption that the transitions are generated by a Gumbel-Max SCM\@.  

\textbf{Target Policy}: The target policy is learned using Policy Iteration on the parameters of the learned MDP\@. Because the target policy is learned using a limited number of samples, as well as an incomplete set of variables, it should perform poorly relative to the behavior policy.

\subsection{Off-Policy Evaluation Can Be Misleading}%
\label{sec:off_policy_evaluation_is_misleading}

\begin{figure}[t]
  \centering
  \includegraphics[width=\linewidth]{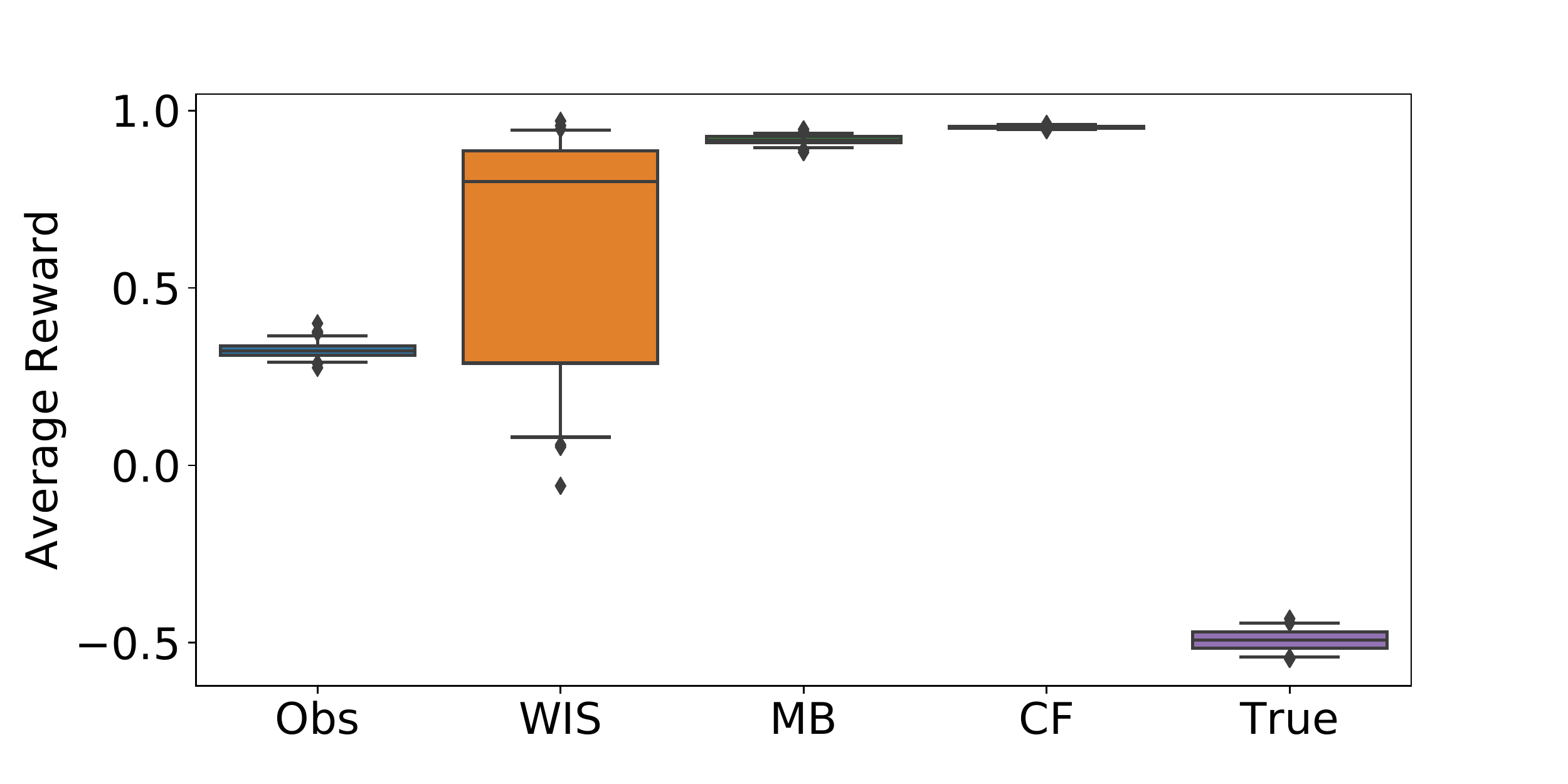}
  \vspace{-8mm}
  \caption{Estimated reward under the target (RL) policy, with 95\% uncertainty intervals generated through 100 bootstrapped samples (with replacement) of the same 1000 observed trajectories (for 1,2,4) and of 1000 new trajectories under the target policy (for 3,5). \textbf{(1) Obs}: Observed reward under the behavior policy.  \textbf{(2) WIS}: Estimated reward under the target policy using weighted importance sampling.  \textbf{(3) MB}: Estimated reward using the learned MDP as a generative model.  \textbf{(4) CF}: Estimated reward over counterfactual trajectories (5 per observed trajectory).  \textbf{(5) True}: Observed reward under the target policy, over 1000 newly simulated trajectories.}%
  \label{fig:ope}
\end{figure}

First, we demonstrate what might be done to evaluate this target policy without the use of counterfactual tools. In Figure~\ref{fig:ope}, we compare the observed reward of the actual trajectories against the estimated reward of the target policy.  Using weighted importance sampling on the given trajectories, the target policy appears superior to the behavior policy.  We also use the parameters of the learned MDP to perform model-based off-policy evaluation (MB-PE), using the MDP as a generative model to simulate trajectories and their expected reward. Both of these suggest that the target policy is superior to the behavior policy.  In reality, the target policy is inferior (as expected by construction), as verified by drawing new samples from the simulator under the target policy.  This corresponds conceptually to what would happen if the target policy were deployed ``in the wild''.

With this in mind, we demonstrate how examining individual counterfactual trajectories gives insight into the target policy.  The first step is to apply counterfactual off-policy evaluation (CF-PE) using the same MDP and the Gumbel-Max SCM\@.  This yields similarly optimistic results as MB-PE\@.  However, by pairing counterfactual outcomes with observed outcomes of individual patients, we can investigate \textit{why} the learned MDP concludes (wrongly) that the target policy would be so successful.  

\subsection{Identification of Informative Trajectories}%
\label{sec:id_informative_trajectories}

\begin{figure}[t]
  \centering
  \includegraphics[width=0.5\linewidth]{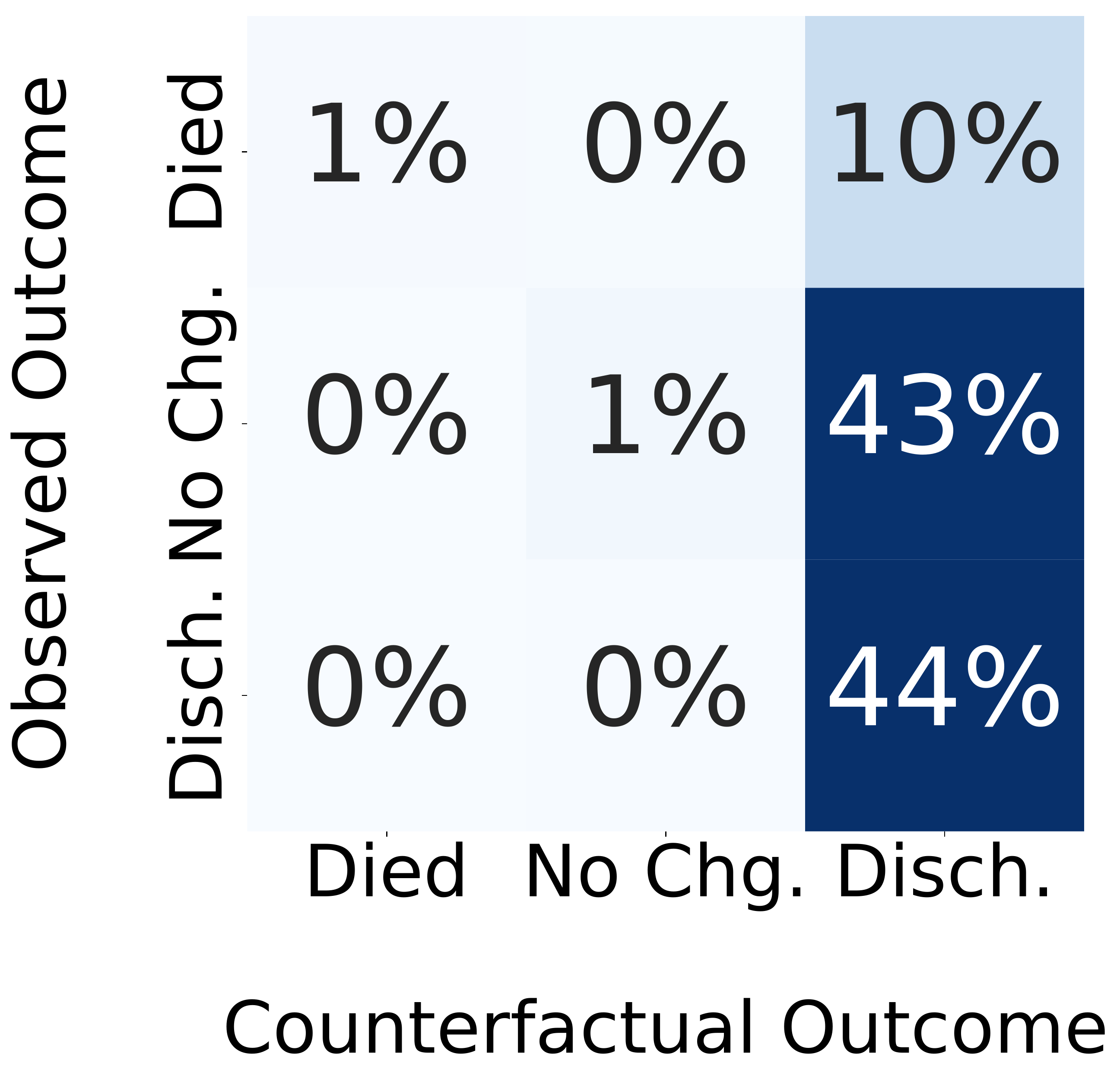}
  \vspace{-2mm}
  \caption{Decomposition of 1000 observed patient trajectories based on observed outcome (Died, no change, and discharged) vs counterfactual outcome under the target policy, using the most common outcome over 5 draws from the counterfactual posterior.}%
  \label{fig:optimistic_decomp}
\end{figure}

To debug this model (without access to a simulator), we can start by drawing counterfactual trajectories for each individual patient under the target policy.  With these in hand, we can assign each individual patient to one of nine categories, based on the most frequently occurring counterfactual outcome (death, no change, or discharge) in Figure~\ref{fig:optimistic_decomp}.  This highlights individual trajectories for further analysis, as discussed in the next section\footnote{We only draw 5 counterfactuals per observed trajectory for illustrative purposes here, but note that standard concentration arguments could be used to quantify how many of these independent draws are required to achieve a desired precision on counterfactual quantities of interest, e.g., the probability of death}.

\subsection{Insights from Examining Individual Trajectories}%
\label{sec:insights_from_examining_individual_trajectories}

Using this decomposition, we can focus on the 10\% of observed trajectories where the model concludes that ``if the physician had applied the target policy, these patients who died would have most likely lived''.

\begin{figure}[t]
  \centering
  \includegraphics[width=0.95\linewidth]{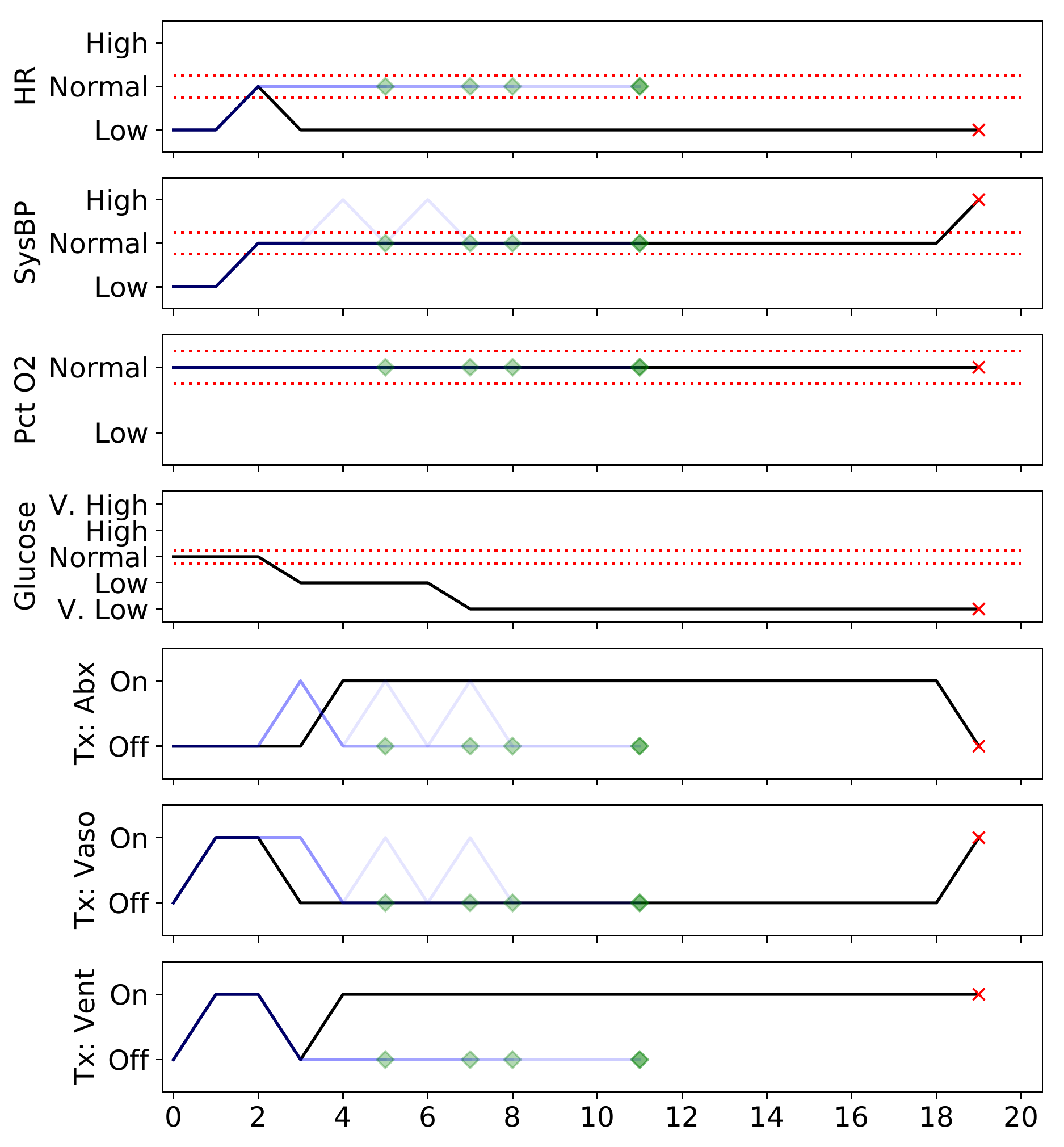}
  \caption{Observed and counterfactual trajectories of a patient.  The first four plots show the progression of vital signs, and the last three show the treatment applied.  For vital signs, the normal range is indicated by red dotted lines.  The black lines show the observed trajectory, which ends in death (signified by the red cross), and the blue lines show five counterfactual trajectories all of which end in discharge, signified by green diamonds. The glucose vital sign was not included in the model, and hence does not have a counterfactual trajectory.  Note how this differs from a newly simulated trajectory of a patient with the same initial state, e.g., all the counterfactual trajectories are identical to the observed trajectory up to a divergence in actions ($t=2$).}%
  \label{fig:debugging_stops_treatment}
\end{figure}

This is a bold statement, but also one that is plausible for domain experts to investigate (e.g., through chart review of these specific patients), to try and understand the rationale.  We illustrate this type of analysis in Figure~\ref{fig:debugging_stops_treatment}, which shows both the observed trajectory and the counterfactual trajectories for a simulated patient.

This example illustrates a dangerous failure mode, where the target policy would have halted treatment despite the glucose vital being dangerously low (e.g., at $t = 5, 7, 8, 11$).  Under the learned MDP, the counterfactual optimistically shows a speedy discharge as a result of halting treatment.  To understand why, recall that discharge occurs when all four vitals are normal and treatment is stopped. Because diabetes and glucose fluctuations are relatively rare, and because the MDP does not observe either, the model learns that there is a high probability of discharge when the first three vitals are normal, and the action of `stop all treatments' is applied.
\section{Conclusion}%
\label{sec:conclusion}

Given the desire to deploy RL policies in high-risk settings (e.g., healthcare), it is important to develop more tools and techniques to introspect these models and the policies they learn.  Our proposed technique can be used to flag individual trajectories for review by a domain expert, based on the fact that they have divergent counterfactual and observed outcomes. Specifically, we gave an example of how this could help to identify sub-optimal (and perhaps dangerous) behavior in a proposed policy.

Applying this approach requires knowing the structural causal model: simply knowing the MDP is insufficient, as demonstrated by our unidentifiability results. The Gumbel-Max SCM is an example of an SCM that may be realistic in some settings. As followup, it would be interesting to understand the sensitivity of counterfactual trajectories to the specific choice of SCM\@.  For instance, while we are not aware of another SCM that both satisfies the counterfactual stability condition and can model any discrete conditional probability distribution, it would be interesting to characterize the space of such alternative SCMs, or prove that no other such SCM exists for $k > 2$ which satisfies these conditions.

\section*{Acknowledgments}%
\label{sec:acknowledgments}

We thank Christina X. Ji and Fredrik D. Johansson for their work on developing an earlier version of the sepsis simulator.  We also thank Fredrik D. Johansson for insightful comments and feedback, as well as Alex D'Amour, Uri Shalit, and David Blei for helpful discussions.  This work was supported in part by Office of Naval Research Award No. N00014-17-1-2791, and Michael was supported in part by the Analog Devices Graduate Fellowship.

\newpage
\appendix
\section*{Appendix}
\setcounter{thmthm}{0}
\setcounter{thmlem}{0}
\setcounter{thmcol}{0}
\section{Omitted Proofs}%
\label{sec:omitted_proofs}

\begin{thmlem}[Counterfactual Decomposition of Expected Reward]
  Let trajectories $\tau$ be drawn from $p^{\pi_{obs}}$.  Let $\tau_{\hat{\pi}}$ be a counterfactual trajectory, drawn from our posterior distribution over the exogenous $U$ variables under the new policy $\hat{\pi}$.  Note that under the SCM, $\tau_{\hat{\pi}}$ is a deterministic function of the exogenous $U$ variables, so we can write $\tau_{\hat{\pi}}(u)$ to be explicit:
\begin{align*}
  &\E_{\hat{\pi}}[ R(\tau) | O_1 = o_1 ] \\
  &= \int_{\tau} p^{\pi_{obs}}(\tau | O_1 = o_1) \E_{u \sim p^{\pi_{obs}}(u | \tau)}[R(\tau_{\hat{\pi}}(u))] d\tau
\end{align*}
\label{thm:decomposition_appendix}
\end{thmlem}%
\begin{proof}
  This proof is similar to the proof of Lemma 1 from \citep{Buesing2018}, but is spelled out here for the sake of clarity.  Recall that the distribution of noise variables $U$ is the same for every intervention / policy.  Thus, $p^{\pi_{obs}}(u) = p^{\hat{\pi}}(u) = p(u)$.  We will write $p'$ and $\hat{p}$ for $p^{\pi_{obs}}$ and $p^{\hat{\pi}}$ respectively to simplify notation.  
  
  Furthermore, recall that all variables are a deterministic function of their parents in the causal DAG implied by the SCM\@.  Most importantly, this means that the trajectory $\tau$ is a deterministic function of the policy $\pi$ and the exogenous variables $U$.  With that in mind, let $\tau_{\hat{\pi}}(u)$ indicate the trajectory $\tau$ as a deterministic function of $\hat{\pi}$ and $u$.  We will occasionally use indicator functions to indicate whether or not a deterministic value is compatible with the variables that determine it, e.g., $\1{\tau | u, \pi}$ is equivalent to the indicator for $\1{\tau = \tau_{\pi}(u)}$.  Note that the first observation is independent of the policy, and is just a function of the exogenous $U$, so we will write $\1{o_1 | u}$ in that case.  For simplicity, we will remove the conditioning on $O_1$ to start with:
\begin{align}
  &\E_{\hat{p}}[ R(\tau)] \nonumber \\ 
  &= \int R(\tau_{\hat{\pi}}(u)) \cdot \hat{p}(u) du \label{eq:a1}\\
  &= \int R(\tau_{\hat{\pi}}(u)) \cdot p'(u) du \label{eq:a3}\\ 
  &= \int R(\tau_{\hat{\pi}}(u)) \cdot \left( \int p'(\tau, u) d\tau \right) du \label{eq:a4}\\ %
  &= \int \int R(\tau_{\hat{\pi}}(u)) \cdot p'(u | \tau) \cdot p'(\tau) du d\tau \label{eq:a5}\\ %
  &= \E_{\tau \sim p'} \left[ \int R(\tau_{\hat{\pi}}(u)) \cdot p'(u | \tau) du \right] \label{eq:a6} \\%
  &= \E_{\tau \sim p'} \E_{u \sim p'(u | \tau)} \left[ R(\tau_{\hat{\pi}}(u)) \right] \label{eq:a8} \\%
  &= \int_\tau p^{\pi_{obs}} (\tau) \E_{u \sim p'(u | \tau)} \left[ R(\tau_{\hat{\pi}}(u)) \right] \label{eq:a9} d\tau%
\end{align}
In step~(\ref{eq:a1}) we are just using the definition of the expectation under $\hat{p}$, along with the notation $\tau_{\hat{\pi}}(u)$ to indicate that the trajectory is a deterministic function of the exogenous $u$ and the policy $\hat{\pi}$.  In step~(\ref{eq:a3}) we replace $\hat{p}(u)$ with $p'(u)$ because they are equivalent, as noted earlier.  In step~(\ref{eq:a4}) we expand $p'(u)$ over possible trajectories $\tau$ arising from the observed policy.  In step~(\ref{eq:a5}) we rearrange terms and swap the order of the integral, and in step~(\ref{eq:a6}) we rewrite the outer integral as an expectation.  In step~(\ref{eq:a8}) we further condense notation, and then expand in step~(\ref{eq:a9}) to match the notation in the Lemma.  If we introduce the conditioning on $O_1$, we see that it is substantively the same.
\begin{align}
  &\E_{\hat{p}}[ R(\tau) | o_1] \nonumber \\ 
  &= \int R(\tau_{\hat{\pi}}(u)) \cdot \1{o_1 | u} \cdot \hat{p}(u) du \label{eq:a9.5}\\
  &= \int R(\tau_{\hat{\pi}}(u)) \cdot \1{o_1 | u} \cdot p'(u) du \label{eq:a10}\\
  &= \int R(\tau_{\hat{\pi}}(u)) \cdot p'(u | o_1) du \label{eq:a13}\\ 
  &= \int R(\tau_{\hat{\pi}}(u)) \cdot \left( \int p'(\tau, u | o_1) d\tau \right) du \label{eq:a14}\\ %
  &= \int \int R(\tau_{\hat{\pi}}(u)) \cdot p'(u | \tau) \cdot p'(\tau | o_1) du d\tau \label{eq:a15}\\ %
  &= \int p'(\tau | o_1) \left[  \int R(\tau_{\hat{\pi}}(u)) \cdot p'(u | \tau) du \right] d\tau \label{eq:a16}\\ %
  &= \int_\tau p'(\tau | o_1) \E_{u \sim p'(u|\tau)} [R(\tau_{\hat{\pi}}(u))] d\tau \label{eq:a17} %
\end{align}
The main difference in this case is that is just that we carry the indicator into the prior on $U$ at step~(\ref{eq:a13}), which we can do because $O_1$ does not depend on the policy that is applied.  Note that Equation~(\ref{eq:a17}) matches the statement of the Lemma.
\end{proof}

\begin{thmcol}[Counterfactual Decomposition of $\delta_o$]
\begin{align*}
  &\delta_{o}\coloneqq \E_{\hat{\pi}}[ R(\tau) | O_1 = o_1 ] - \E_{obs} [ R(\tau) | O_1 = o_1 ] \\
             &= \int_{\tau} p^{\pi_{obs}}(\tau | O_1 = o_1) \E_{u \sim p^{\pi_{obs}}(u | \tau)}[R(\tau_{\hat{\pi}}(u)) - R(\tau)]d\tau
\end{align*}%
\label{thm:decomposition_col_appendix}
\end{thmcol}
\begin{proof}
  By Lemma~\ref{thm:decomposition_appendix}, we have it that
  \begin{align*}
  &\delta_{o} \coloneqq \E_{\hat{\pi}}[ R(\tau) | O_1 = o ] - \E_{obs} [ R(\tau) | O_1 = o ] \\
  &= \int_\tau p'(\tau | o_1) \E_{u \sim p'(u|\tau)} [R(\tau_{\hat{\pi}}(u))] d\tau \\
  &\ \ - \int_\tau p'(\tau | o_1) \E_{u \sim p'(u|\tau)} [R(\tau_{\pi_{obs}}(u))] d\tau \\
             &= \int_{\tau} p^{\pi_{obs}}(\tau | O_1 = o_1) \E_{u \sim p^{\pi_{obs}}(u | \tau)}[R(\tau_{\hat{\pi}}(u)) - R(\tau)]d\tau
  \end{align*}
  Note that in the last step, we recognize that $\P_{u \sim p'(u | \tau)}[\tau_{\pi_{obs}}(u) = \tau] = 1$, because the posterior density over $u$ is zero for all $u$ such that $\tau_{\pi_{obs}}(u) \neq \tau$.
\end{proof}

\newpage

\begin{thmthm}
  Let $Y = f_y(t, u)$ be the SCM for a binary variable $Y$, where $T$ is also a binary variable.  If this SCM satisfies the counterfactual stability property, then it also satisfies the monotonicity property with respect to $T$.
\end{thmthm}

\begin{proof}
  We collect Definitions~(4) and (5) here for ease of reference
  
  \textbf{Monotonicity:} An SCM of a binary variable $Y$ is monotonic relative to a binary variable $T$ if and only if it has the following property: $\E[Y | do(T = t)] \geq \E[Y | do(T = t')] \implies f_y(t, u) \geq f_y(t', u), \ \forall u$.  We can write equivalently that the following event never occurs, in the case where $\E[Y | do(T = 1)] \geq \E[Y | do(T = 0)]$: $Y_{do(T = 1)} = 0 \land Y_{do(T = 0)} = 1$.  Conversely for $\E[Y | do(T = 1)] \leq \E[Y | do(T = 0)]$, the following event never occurs: $Y_{do(T = 1)} = 1 \land Y_{do(T = 0)} = 0$
 
  \textbf{Counterfactual Stability}: An SCM of a categorical variable $Y$ satisfies \textit{counterfactual stability} if it has the following property:  If we observe $Y_{I} = i$, then for all $j \neq i$, the condition $\frac{p'_i}{p_i} \geq \frac{p'_j}{p_j}$ implies that $P^{\cM|Y_I=i; I'}(Y = j) = 0$.  That is, if we observed $Y = i$ under intervention $I$, then the counterfactual outcome under $I'$ cannot be equal to $Y = j$ unless the multiplicative change in $p_i$ is less than the multiplicative change in $p_j$

  To simplify notation further, let $p^{t=1} \coloneqq P(Y = 1 | do(T = 1))$, $p^{t=0} \coloneqq P(Y = 1 | do(T = 0))$, and let $Y_t \coloneqq Y_{do(T = t)}$.  Without loss of generality, assume that $p^{t=1} \geq p^{t=0}$.
  
  To show that counterfactual stability implies monotonicity, we want to show that the probability of the event $(Y_1 = 0 \land Y_0 = 1)$ is equal to zero.  We will do so by proving both cases: First that $P^{\cM|Y_0 = 1;do(T = 1)}(Y = 0) = 0$ and second that $P^{\cM|Y_1 = 0;do(T = 0)}(Y = 1) = 0$.  We can start with the assumption that $p^{t=1} \geq p^{t=0}$ and write:
  \begin{align*}
    p^{t=1} &\geq p^{t=0} \\
    \implies p^{t=1} (1 - p^{t=0}) &\geq p^{t=0} (1 - p^{t=1}) \\
    \implies \frac{p^{t=1}}{p^{t=0}} &\geq \frac{(1 - p^{t=1})}{(1 - p^{t=0})}
  \end{align*}
  
  Using the counterfactual stability condition, the last inequality implies that if we observe $Y_0 = 1$, then the counterfactual probability of $Y_1 = 0$ is equal to $P^{\cM | Y_0 = 1; do(T = 1)}(Y = 0) = 0$, as desired.  For the second case, where we observe $Y_1 = 0$, we can simply manipulate the inequality to see that
  \[\frac{(1 - p^{t=0})}{(1 - p^{t=1})} \geq \frac{p^{t=0}}{p^{t=1}} \]
  Which yields the conclusion that $P^{\cM | Y_1 = 0; do(T = 0)}(Y = 1) = 0$, as desired, completing the proof.
\end{proof}

\begin{thmthm}
  The Gumbel-Max SCM satisfies the counterfactual stability condition.%
\end{thmthm}
\begin{proof}
  Recall that we write the shorthand $p_i \coloneqq P^{\cM;I}(Y = i)$, and $p_i' \coloneqq P^{\cM; I'}(Y = i)$.  Suppose that $Y$ is generated from a Gumbel-Max SCM $\cM$ under intervention $I$, and we observe that $Y_I = i$.  The Gumbel-Max SCM implies that almost surely:
  \begin{align}
    \log p_i + g^{(i)} &> \log p_j + g^{(j)}\ \ \forall j \neq i \label{eq:appendix-1}
  \end{align}
  To demonstrate that the Gumbel-Max SCM satisfies the counterfactual stability condition, we need to demonstrate that $\frac{p'_i}{p_i} \geq \frac{p'_j}{p_j} \implies P^{\cM| Y_I = i; I'}(Y = j) = 0$ for all $j \neq i$.  
  
  We will proceed by proving the contrapositive, that for all $j \neq i$, $P^{\cM| Y_I = i; I'}(Y = j) \neq 0 \implies \frac{p'_i}{p_i} < \frac{p'_j}{p_j}$.

  Fix some index $j \neq i$.  The condition $P^{\cM| Y_I = i; I'}(Y = j) \neq 0$ implies that there exist values $g^{(i)}, g^{(j)}$ such that 
  \begin{align}
    \log p'_i + g^{(i)} &< \log p'_j + g^{(j)} \label{eq:appendix-1b}
  \end{align}
  
  Because the Gumbel variables $g^{(i)}, g^{(j)}$ are fixed across interventions, this implies there exist values for these variables which satisfy both inequalities~(\ref{eq:appendix-1}) and~(\ref{eq:appendix-1b}).  Thus, we proceed by subtracting inequality~(\ref{eq:appendix-1}) from inequality~(\ref{eq:appendix-1b}), maintaining the direction of the inequality and cancelling out the Gumbel terms.  The rest is straightforward manipulation using the monotonicity of the logarithm.
  \begin{align*}
    \log p'_i - \log p_i &< \log p'_j - \log p_j \\
    \log (p'_i / p_i) &< \log (p'_j / p_j) \\
    (p'_i / p_i) &< (p'_j / p_j)
  \end{align*}
 This demonstrates that $P^{\cM|Y_I = i; I'}(Y = j) \neq 0 \implies (p'_i / p_i) < (p'_j / p_j)$ as desired, and taking the contrapositive completes the proof. 
\end{proof}

\section{Non-Identifiability Example}%
\label{sec:non_identifiability_example}

\begin{figure}[h]
  \centering
  \includegraphics[width=0.95\linewidth]{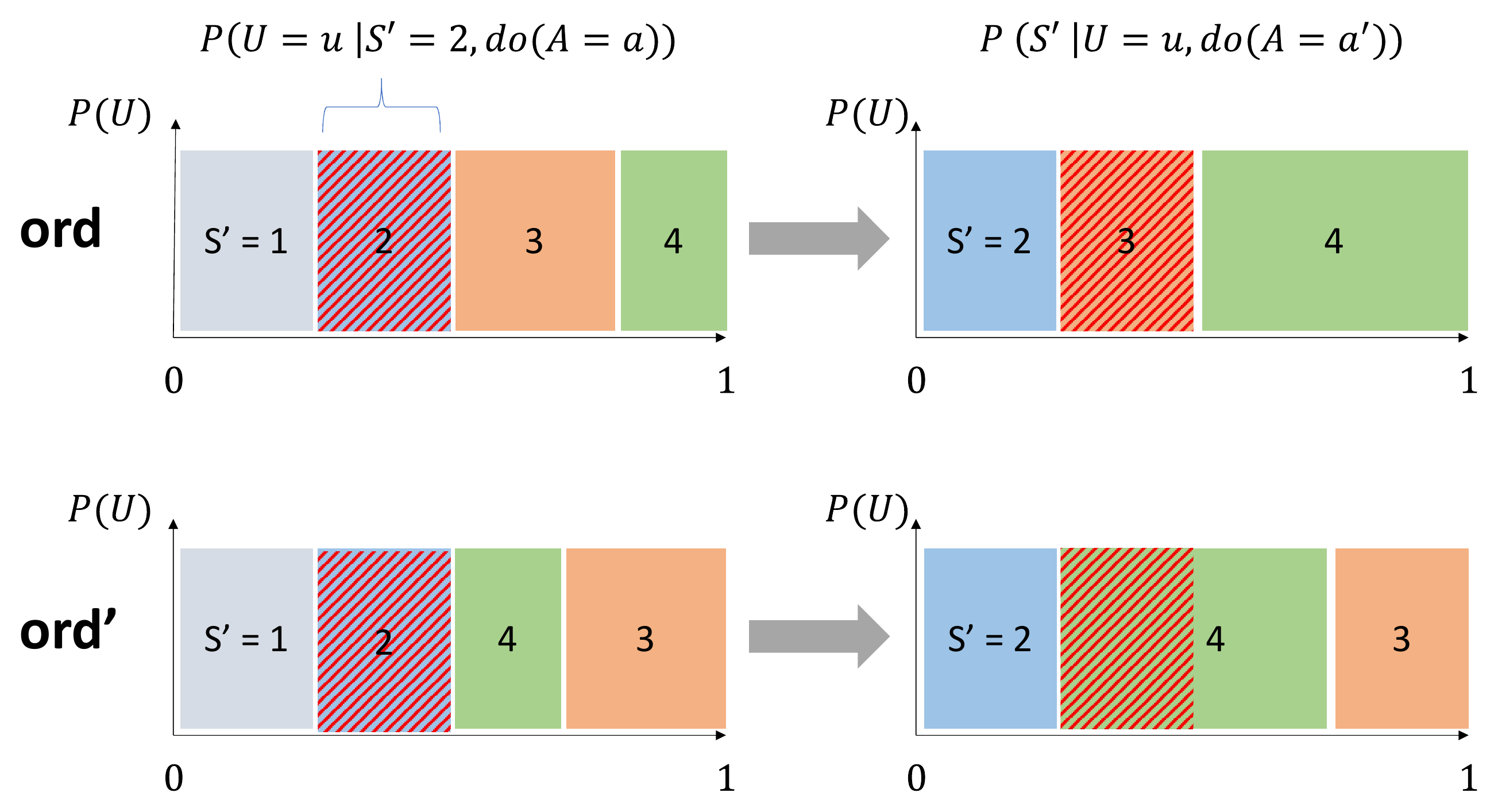}
  \caption{Example of non-identifiability of categorical counterfactual outcomes.  The orderings \textbf{ord} and \textbf{ord'} both define a causal mechanism $S' = f(S, A, U)$ with $U \sim Unif(0, 1)$ that replicates the interventional probability distribution $P(S' | S, A)$.  On the left-hand side, the red shading represents the posterior $P(U | S' = 2, A = a, S = s)$, and when this posterior is used on the right-hand side to sample from the counterfactual distribution, these ordering produce different counterfactual outcomes ($S' = 3$ in the case of \textbf{ord} and $S' = 4$ in the case of \textbf{ord'})}%
  \label{fig:nonid_example}
\end{figure}

Figure~\ref{fig:nonid_example} gives a visual depiction of the unidentifiability example given in Section 3.1.

\section{Experimental Details}%
\label{sec:experimental_details}

\subsection{Sepsis Simulator}%
\label{sub:sepsis_simulator}

All the code required to reproduce our experiments (including the figures in this appendix) is available online at \url{https://www.github.com/clinicalml/gumbel-max-scm}, and we refer to that for more in-depth information about our simulator setup.

\subsection{Impact of hidden state}%
\label{sub:impact_of_hidden_state}

In the experiments given in the paper, we hide the glucose and diabetes state from the model of dynamics used for the RL policy.  In this section we explore the impact of that choice on the off-policy evaluations used in the paper, as well as on the quality of the RL policy.

\begin{figure}[t]
  \centering
  \includegraphics[width=0.95\linewidth]{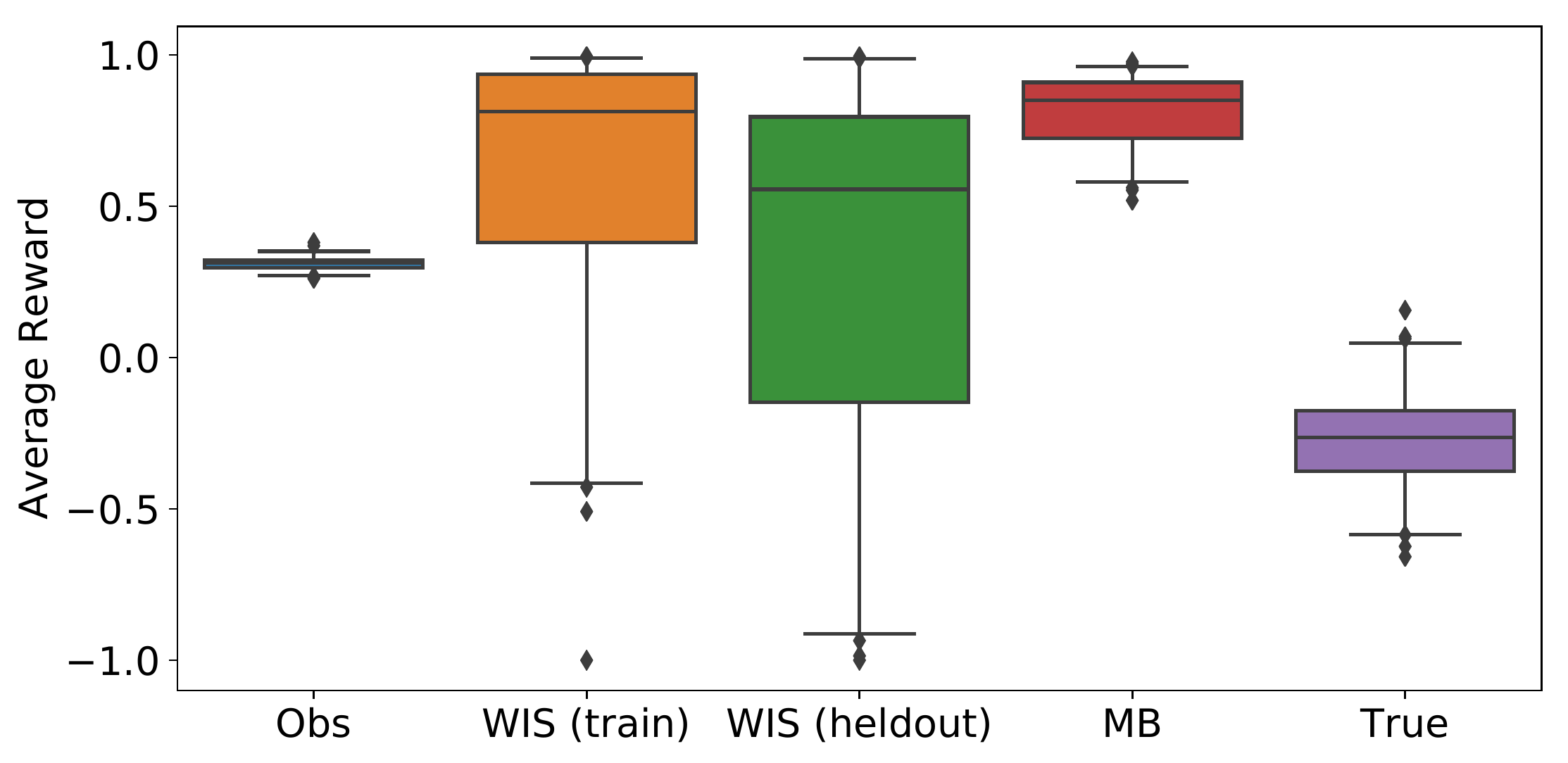}
  \caption{Boxplots show the median and intervals which capture 95\% of the 100 evaluations, each time with a newly simulated set of 1000 episodes used for training and 1000 episodes used for the held-out WIS estimator;  WIS (train) is used on the training episodes, as in the main paper, and WIS (held-out) is performed on the held-out set of 1000 episodes}%
  \label{fig:heldout-1k}
\end{figure}

To demonstrate, in Figure~\ref{fig:heldout-1k}, we replicate Figure 3 from the main paper, but with some important differences.  First, instead of using 100 bootstrapped samples of the original 1000 trajectories, we instead repeat the entire process 100 times, with an independent set of trajectories drawn from the simulator in each case.  These uncertainty intervals are wider, reflecting the variation which is not captured by bootstrapping alone.  Second, we compare the use of a WIS estimator used on the training data (i.e., the original 1000 episodes used to learn the model of dynamics), with a WIS estimator used on a held-out set of 1000 independent episodes.  While the example given in the paper is meant to conceptually capture what might happen in a single analysis (where only a single set of trajectories is available), Figure~\ref{fig:heldout-1k} demonstrates the variability across analyses, including those with access to a large held-out set of trajectories.

\begin{figure}[t]
  \centering
  \includegraphics[width=0.95\linewidth]{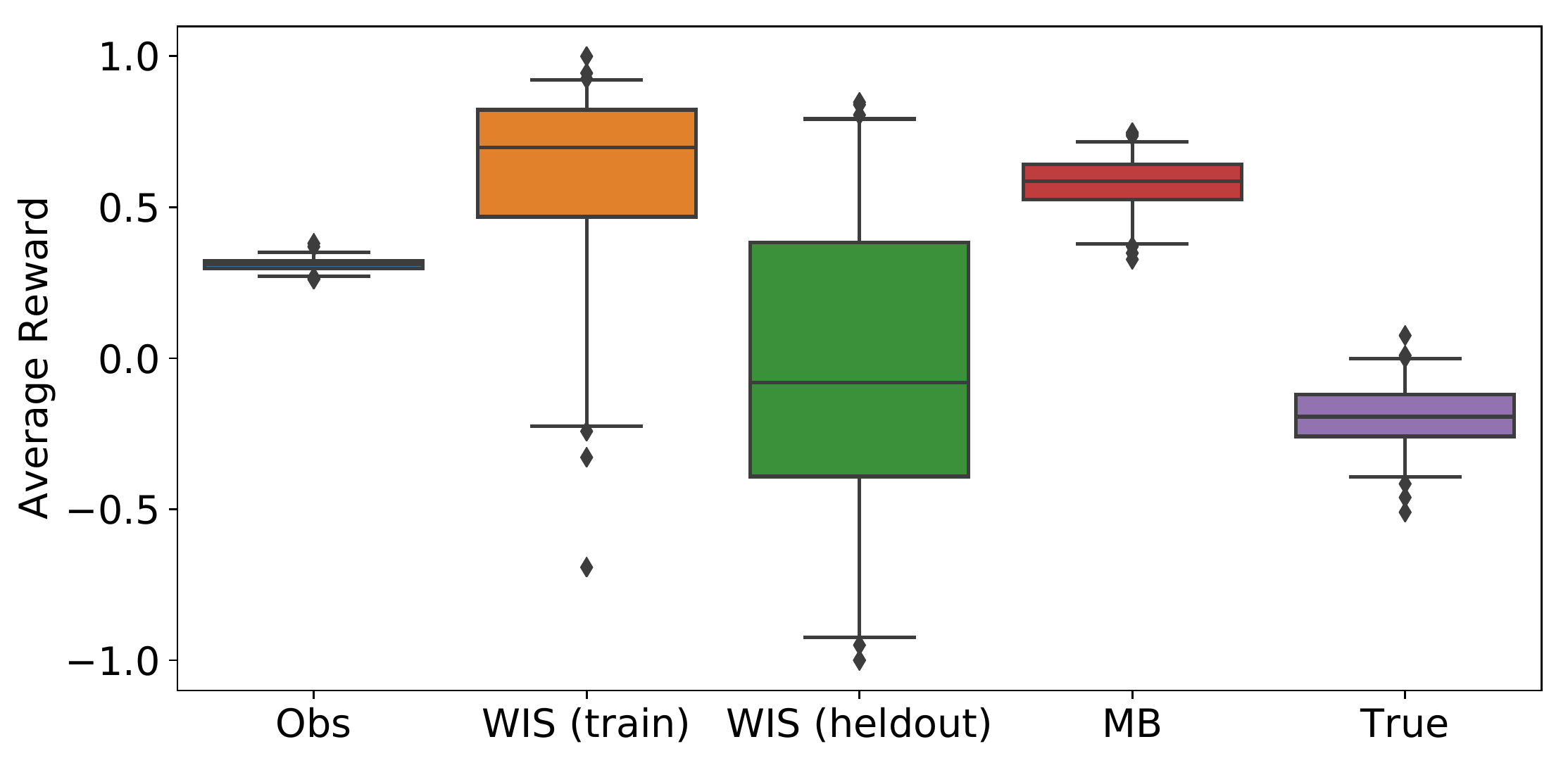}
  \caption{Same setup as Figure~\ref{fig:heldout-1k}, but allowing the model of dynamics used by the MDP to see the full state}%
  \label{fig:heldout-full-state-1k}
\end{figure}

Towards understanding the impact of hiding variables from the RL policy, we performed the same experiment again, but giving the RL policy access to the entire state space.  The results are shown in Figure~\ref{fig:heldout-full-state-1k}, and the results from both figures are shown in Table~\ref{tab:full-results}

\begin{table}[h]
  \centering
  \caption{Performance given as Mean (95\% CI) from Figures~\ref{fig:heldout-1k}-~\ref{fig:heldout-full-state-1k}}%
  \label{tab:full-results}
\begin{tabular}{l l l}
  \centering
  & Hidden state & No hidden state\\
  \toprule
  Observed Reward &0.31 (0.27, 0.35) & 0.31 (0.27, 0.35)\\
  WIS (train) &0.61 (-0.42, 0.99) & 0.58 (-0.23, 0.92)\\
  WIS (heldout) &0.32 (-0.92, 0.99)& -0.04 (-0.94, 0.80)\\
  MB Estimate&0.81 (0.57, 0.96) & 0.58 (0.37, 0.73)\\
  True RL Reward&-0.27 (-0.59, 0.05) & -0.19 (-0.41, 0.00)\\
  \bottomrule
\end{tabular}
\end{table}

There are several reasons why weighted importance sampling, and other off-policy evaluation methods, could fail to capture the true performance of a target policy.  These include issues like confounding and small sample sizes, as discussed in \citep{Gottesman2019}.  In this particular synthetic example, all of the following factors may play a role in the above results, but it is difficult to say conclusively how strong each factor is, and how they interact to produce the results: (i) Confounding due to unobserved states, (ii) sample complexity of learning the MDP, which is more pronounced when all state information is observed (144 states vs 1440 states), and (iii) small sample sizes in both the training and held-out datasets.

With that in mind, we believe that building a more comprehensive simulated environment, in which these various factors can be disentangled more precisely, would be a valuable direction for future work.  In addition, we believe such an environment would be useful for evaluation of a variety of off-policy techniques beyond the limited set discussed in the paper e.g., more recently developed methods such as~\cite{Thomas2016, NIPS2018_7530}.

\newpage

\bibliographystyle{icml2019}
\bibliography{full}

\end{document}